\documentclass[11pt,a4paper]{article}
\usepackage[hyperref]{naaclhlt2019}
\usepackage{times}
\usepackage{latexsym}

\usepackage{caption}
\usepackage{subcaption}
\usepackage{threeparttable}
\usepackage{multirow, array}
\usepackage{booktabs}
\usepackage{amsmath}
\usepackage{graphicx}
\usepackage{textcomp}

\usepackage{url}

\aclfinalcopy 

\title{Acquiring Annotated Data with Cross-lingual Explicitation \\ for Implicit Discourse Relation Classification}

\author{Wei Shi$^\dag$, ~ Frances Yung$^\dag$ \and~Vera Demberg$^{\dag,\ddag}$\\
	$^\dag$Dept. of Language Science and Technology\\ 
	$^\ddag$Dept. of Mathematics and Computer Science, Saarland University\\
    Saarland Informatic Campus, 66123 Saarbr\"ucken, Germany\\
	{\tt\{w.shi, frances, vera\}@coli.uni-saarland.de}}

\date{}

\begin{document}
\maketitle
\begin{abstract}
  Implicit discourse relation classification is one of the most challenging and important tasks in discourse parsing, due to the lack of connectives as strong linguistic cues. 
A principle bottleneck to further improvement is the shortage of training data (ca.~18k instances in the Penn Discourse Treebank (PDTB)). \citet{shi2017using} proposed to acquire additional data by exploiting connectives in translation: human translators mark discourse relations which are implicit in the source language explicitly in the translation. Using back-translations of such explicitated connectives improves discourse relation parsing performance. This paper addresses the open question of whether the choice of the translation language matters, and whether multiple translations into different languages can be effectively used to improve the quality of the additional data. 
\end{abstract}

\section{Introduction}

Discourse relations connect two sentences/clauses to each other. The identification of discourse relations is an important step in natural language understanding and is beneficial to various downstream NLP applications such as text summarization \cite{yoshida2014dependency,gerani2014abstractive}, question answering \cite{verberne2007evaluating,jansen2014discourse}, machine translation \cite{guzman2014using,meyer2015disambiguating}, and so on. 

Discourse relations can be marked explicitly using a discourse connective or discourse adverbial such as  ``because'', ``but'', ``however'', see example \ref{eg_1}. Explicitly marked relations are relatively easy to classify automatically \citep{pitler2008easily}. In example \ref{eg_2}, the causal relation is not marked explicitly, and can only be inferred from the texts. This second type of case is empirically even more common than explicitly marked relations \citep{prasad2008penn}, but is much harder to classify automatically.  

\begin{enumerate}
\item{\label{eg_1}
  \emph{[No one has worked out the players' average age.]$_{Arg1}$ \underline{\textbf{But}} [most appear to be in their late 30s.]$_{Arg2}$}
	\vspace{-5pt}

	~~~\hfill --- Explicit, Comparison.Contrast  	
    }
\vspace{-5pt}
\item{\label{eg_2}
   \emph{[I want to add one more truck.]$_{Arg1}$ (\textbf{\underline{Implicit = Because}}) 
     [I sense that the business will continue grow.]$_{Arg2}$}
     \vspace{-3pt}
     
     ~~~\hfill	--- Implicit, Contingency.Cause	
    }
\end{enumerate}

The difficulty in classifying implicit discourse relations stems from the lack of strong indicative cues. Early work has already shown that implicit relations cannot be learned from explicit ones by just removing the discourse markers, which may lead to a meaning shift in the examples \cite{sporleder2008using}, making human-annotated relations currently the only reliable source for training implicit discourse relation classification. 


Due to the limited size of available training data, several approaches have been proposed for acquiring additional training data using automatic methods \citep{wang2012implicit,rutherford2015improving}. The most promising approach so far, \citet{shi2017using}, exploits the fact that human translators sometimes insert a connective in their translation even when a relation was implicit in the original text. Using a back-translation method, \citeauthor{shi2017using}~showed that such instances can be used for acquiring additional labeled text. 

\citet{shi2017using} however only used a single target langauge (French), and had no control over the quality of the labels extracted from back-translated connectives. In this paper, we therefore systematically compare the contribution of three target translation languages from different language families: French (a Romance language), German (from the Germanic language family) and Czech (a Slavic language). As all three of these languages are part of the EuroParl corpus, this also allows us to directly test whether higher quality can be achieved by using those instances that were consistently explicitated in several languages. We use cross-lingual explicitation to acquire more reliable implicit discourse relation instances with separate arguments that are from adjacent sentences in a document, and conducted experiments on PDTB benchmark with multiple conventional settings including cross validation. The experimental results show that the performance has been improved significantly with the additional training data, compared with the baseline systems.

\begin{figure*}[!t]
\centering
	\includegraphics[width=0.95\linewidth]{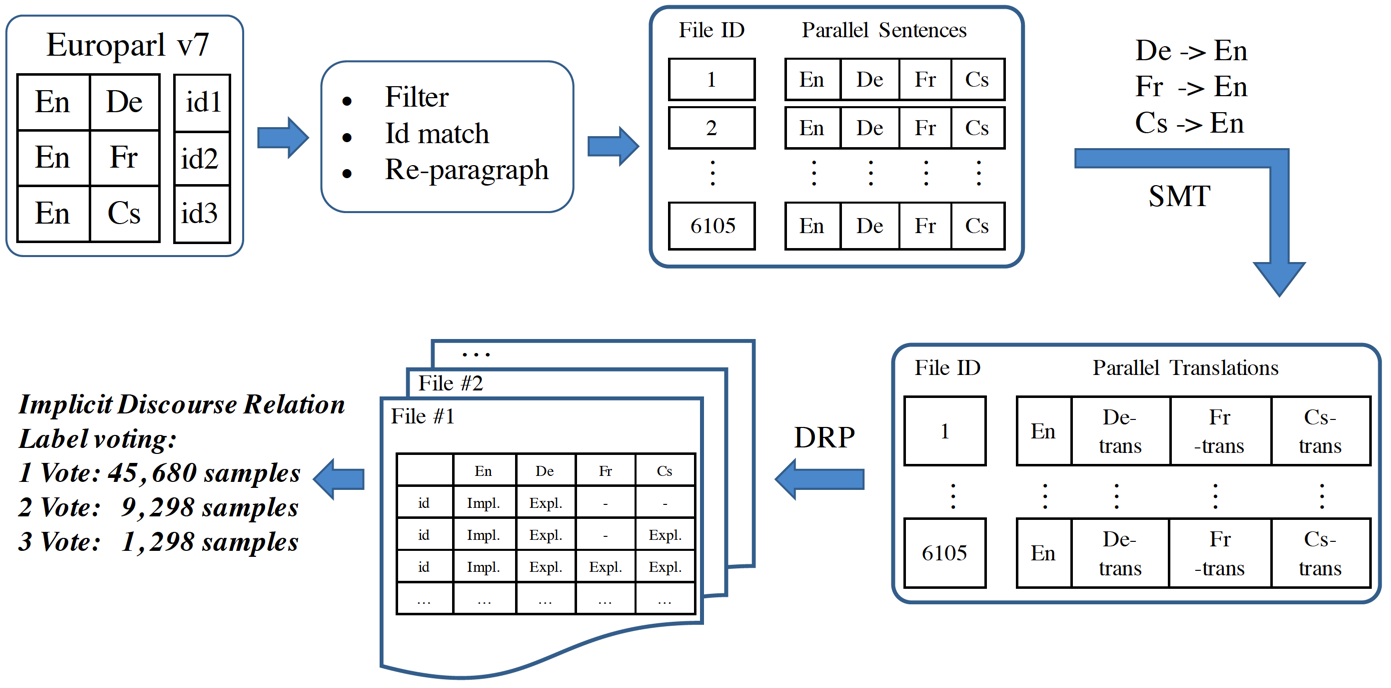}
    \caption{The pipeline of proposed method. ``SMT" and ``DRP" denote statistical machine translation and discourse relation parser respectively.}
    \label{model}
\end{figure*}

\section{Related Work}

Recognizing implicit discourse relation, as one of the most important and challenging part of discourse parser system, has drawn a lot of attention in recent years after the release of PDTB \cite{prasad2008penn}, the largest available corpus with annotated implicit examples, including two shared task in CoNLL-2015 and CoNLL-2016 \cite{xue2015conll,xue2016conll}. 

Early attempts focused on statistical machine learning solutions with sparse linguistic features and linear models. They used several linguistically informed features like polarity tags, Levin verb classes and brown cluster etc. \cite{pitler2009automatic,park2012improving,rutherford2014discovering}. 

Recent methods for discourse relation classification have increasingly relied on neural network architectures \cite{ji2016latent,qin2016implicit,qin2017adversarial,shi2018learning}. However, with the high number of parameters to be trained in more and more complicated deep neural network architectures, the demand for more reliable annotated data has become even more urgent. Data extension has been a longstanding goal in implicit discourse relation classification. \citet{wang2012implicit} proposed to differentiate typical and atypical examples for each relation and augment training data for implicit only by typical explicits. \citet{rutherford2015improving} designed criteria for selecting explicit samples in which connectives can be omitted without changing the interpretation of the discourse. More recently, \citet{shi2017using} proposed a pipeline to automatically label English implicit discourse samples based on explicitation of discourse connectives during human translating in parallel corpora, and achieve substantial improvements in classification. Our work here directly extends theirs by employing document-aligned cross-lingual parallel corpora and majority votes to get more reliable and in-topic annotated implicit discourse relation instances.

\section{Methodology}
Our goal here aims at sentence pairs in cross-lingual corpora where connectives have been inserted by human translators during translating from English to several other languages. After back-translating from other languages to English, explicit relations can be easily identified by discourse parser and then original English sentences would be labeled accordingly. 

We follow the pipeline proposed in \citet{shi2017using}, as illustrated in Figure \ref{model}, with the following differences: 

\begin{itemize}
\vspace{-5pt}
    \item \citet{shi2017using} suffered from the fact that typical sentence-aligned corpora may have some sentences removed and make the sentences no longer coherent to get inter-sentential discourse relation instances. Here we filter and re-paragraph the line-aligned corpus to parallel document-aligned files, which makes it possible to obtain in-topic inter-sentential instances.  After preprocessing, we got 532,542 parallel sentence pairs in 6,105 documents.
    \vspace{-5pt}
    \item  \citet{shi2017using} pointed out that having correct translation of explicit discourse connective is more important than having the correct translation of the whole sentence. In this paper we use a statistical machine translation system instead of a neural one for more stable translations of discourse connectives. 
    \vspace{-5pt}
    \item Instead of a single language pair, we use three language pairs and majority votes between them to get annotated implicit discourse relation instances with high confidence.
\vspace{-5pt}
\end{itemize}

Figure \ref{model} illustrates the pipeline of our approach. It consists of a few steps including preprocessing, back-translating, discourse parsing and majority voting. For each document, we back-translate its German, French and Czech translation back to English with the MT system and parse them with discourse parser. In this way, we can easily identify those instances that are originally implicit but explicit in German, French or Czech. With majority vote by the explicit examples in those three languages, the original English instance could be labeled with different confidences.

\subsection{Preprocessing}
We use European Parliament Proceedings Parallel Corpus (Europarl\footnote{Data is downloaded from \url{http://opus.nlpl.eu/Europarl.php}}) \cite{koehn2005europarl} and choose English-French, German and Czech pairs as our parallel corpora. Each source-target pair consists of source and target sentences along with a sentence ID with which we could easily identify the location of the sentence in certain paragraphs. In order to get document-aligned parallel sentences among all these four languages, we do preprocessing steps as follows:
\begin{itemize}
\item Filtering: remove those sentences that don't have all the three translations in French, German or Czech.

\item ID matching: re-group each sentence into different documents by the sentence IDs.

\item Re-paragraph: rank the sentences in each documents by the ID and re-paragraph them.
\end{itemize}

\subsection{Machine Translation}
We train three MT systems to back-translate French, German and Czech to English. To have word alignments, better and stable back-translations, we employ a statistical machine translation system \textsc{Moses} \cite{koehn2007moses}, trained on the same parallel corpora. Source and target sentences are first tokenized, true-cased and then fed into the system for training. In our case, the translation target texts are identical with the training set of the translation systems; this would not be a problem because our only objective in the translation is to back-translate connectives in the translation into English. On the training set, the translation system achieves BLEU scores of 66.20 (French), 65.30 (German) and 69.05 (Czech).


\subsection{Discourse Parser}
We employ the PDTB-style parser proposed in \cite{lin2014pdtb}, which achieved about 96\% accuracy on explicit connective identification, to pick up those explicit examples in back-translations in each document. Following the definitions of discourse relations in the PDTB that the arguments of the implicit discourse relations should be adjacent sentences but not for the explicit relations, we screen out all those explicit samples from the outputs of the parser that don't have consecutive arguments.

\subsection{Majority Vote}
After parsing the back-translations of French, German and Czech, we can compare whether they contain explicit relations which connect the same relational arguments. The analysis of this subset then allows us to identify those instances that could be labeled with high confidence, i.e.~where back-translations from all three languages allow us to infer the same coherence label. Note that it is not necessarily the case that all back-translations contain an explicitation for the same instance (for instance, the French translator may have explicitated a relation, while the German and the Czech translators didn't do so), or that they propose \textit{the same} coherence label: the human translation can introduce ``noise'' in the sense of the human translators inferring different coherence relations, the machine translation model can introduce errors in back-translation, and the discourse parser can mislabel ambiguous explicit connectives. When we use back-translations of several languages, the idea is that we can eliminate much of this noise by selecting only those instances where all back-translations agree with one another, or the ones where at least two back-translations allow us to infer identical labels.

\begin{figure}[!t]
    \centering
    \includegraphics[width=0.95\linewidth]{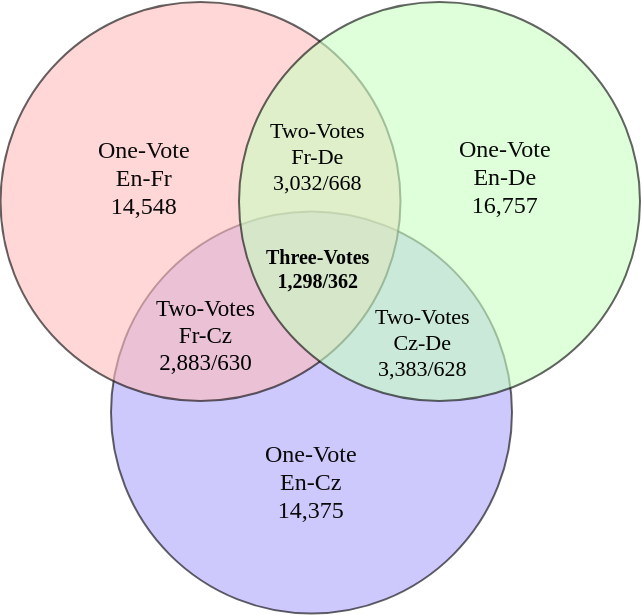}
    \caption{Numbers of implicit discourse relation instances from different agreements of explicit instances in three back-translations. En-Fr denotes instances that are implicit in English but explicit in back-translation of French, same for En-De and En-Cz. The overlap means they share the same relational arguments. The numbers under ``Two-Votes" and ``Three-Votes" are the numbers of discourse relation agreement / disagreement between explicits in back-translations of two or three languages. }
    \label{fig:number_vote}
\end{figure}

Figure \ref{fig:number_vote} illustrates the number of automatically labeled implicit discourse relation examples together with the information of how many of the instances that just one, two or all three back-translations provided the same labels.

In the One Vote agreement, every explicit relation has been accepted and the original implicit English sentences have been annotated correspondingly. Likewise, Two Votes agreement needs at least two out of three languages to have the same explicit relation label after back-translation; agreement between all three back-translations is denoted as Three Votes.

\begin{figure}[!t]
\centering
	\includegraphics[width=0.98\linewidth]{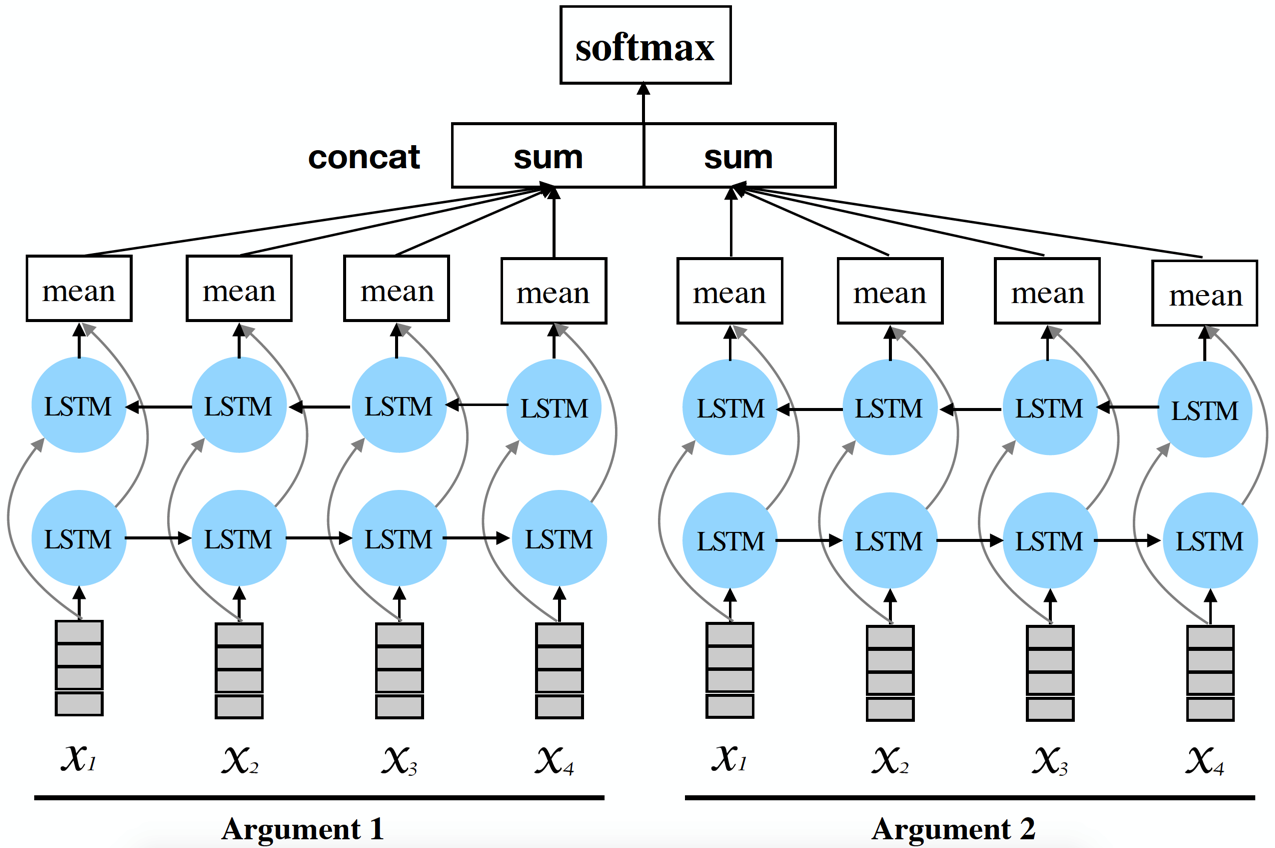}
    \caption{Bi-LSTM network for implicit discoure relation classification.}
    \label{fig:bi_lstm}
\end{figure}


\section{Experiments and Results}
\subsection{Data}
\noindent \textbf{Europarl Corpora:} The parallel corpora used here are from Europarl \cite{koehn2005europarl}, it contains about 2.05M English-French, 1.96M English-German and 0.65M English-Czech pairs. After preprocessing, we got about 0.53M parallel sentence pairs in all these four languages.

\noindent \textbf{The Penn Discourse Treebank (PDTB):} PDTB \cite{prasad2008penn} is the largest available manually annotated corpus of discourse relations from Wall Street Journal. Each discourse relation has been annotated in three hierarchy levels. In this paper, we follow the previous conventional settings and focus on the second-level 11-ways classification \cite{lin2009recognizing, ji2015one, rutherford2017systematic, shi2017using}, after removing the relations with few instances.




\subsection{Implicit discourse relation classification}
To evaluate whether the extracted data is helpful to this task, we use a simple and effective bidirectional Long Short-Term Memory (LSTM) \cite{hochreiter1997long} network. 

A LSTM recurrent neural network processes a variable-length sequence $x=(x_1, x_2, ..., x_n)$. At time step $t$, the state of memory cell $c_t$ and hidden $h_t$ are calculated with the Equations \ref{lstm}:

\begin{equation}
    \label{lstm}
    \small
    \begin{gathered}
        \left[\begin{array}{c} i_t \\ f_t \\ o_t \\ \hat{c_t} \end{array} \right] 
        = \left[\begin{array}{c} \sigma \\ \sigma \\ \sigma \\ \tanh \end{array} \right] W \cdot [h_{t-1}, x_t] 
        \\
c_t = f_t \odot c_{t-1} + i_t \odot \hat{c_t}
\\
h_t = o_t \odot \tanh(c_t)\\
\end{gathered}
\end{equation}

After being mapped to vectors, words are fed into the network sequentially. Hidden states of LSTM cell from different directions are averaged. The representations of two arguments from two separate bi-LSTMs are concatenated before being fed into a softmax layer for prediction. The architecture is illustrated in Figure \ref{fig:bi_lstm}.

\noindent \textbf{Implementation:} The model is implemented in Pytorch\footnote{\url{https://pytorch.org/}}. All the parameters are initialized uniformly at random. We employ cross-entropy as our cost function, Adagrad with learning rate of 0.01 as the optimization algorithm and set the dropout layers after embedding and output layer with drop rates of 0.5 and 0.2 respectively. The word vectors are pre-trained word embeddings from Word2Vec\footnote{\url{https://code.google.com/archive/p/word2vec/}}.

\noindent \textbf{Settings:} We follow the previous works and evaluate our data on second-level 11-ways classification on PDTB with 3 settings: \citet{lin2009recognizing} (denotes as PDTB-Lin) uses sections 2-21, 22 and 23 as train, dev and test set; \citet{ji2015one} uses sections 2-20, 0-1 and 21-22 as train, dev and test set; Moreover, we also use 10-folds cross validation among sections 0-23 \cite{shi2017need}. For each experiment, the additional data is only added into the training set.

\begin{figure}[!ht]
\centering
	\includegraphics[width=0.95\linewidth]{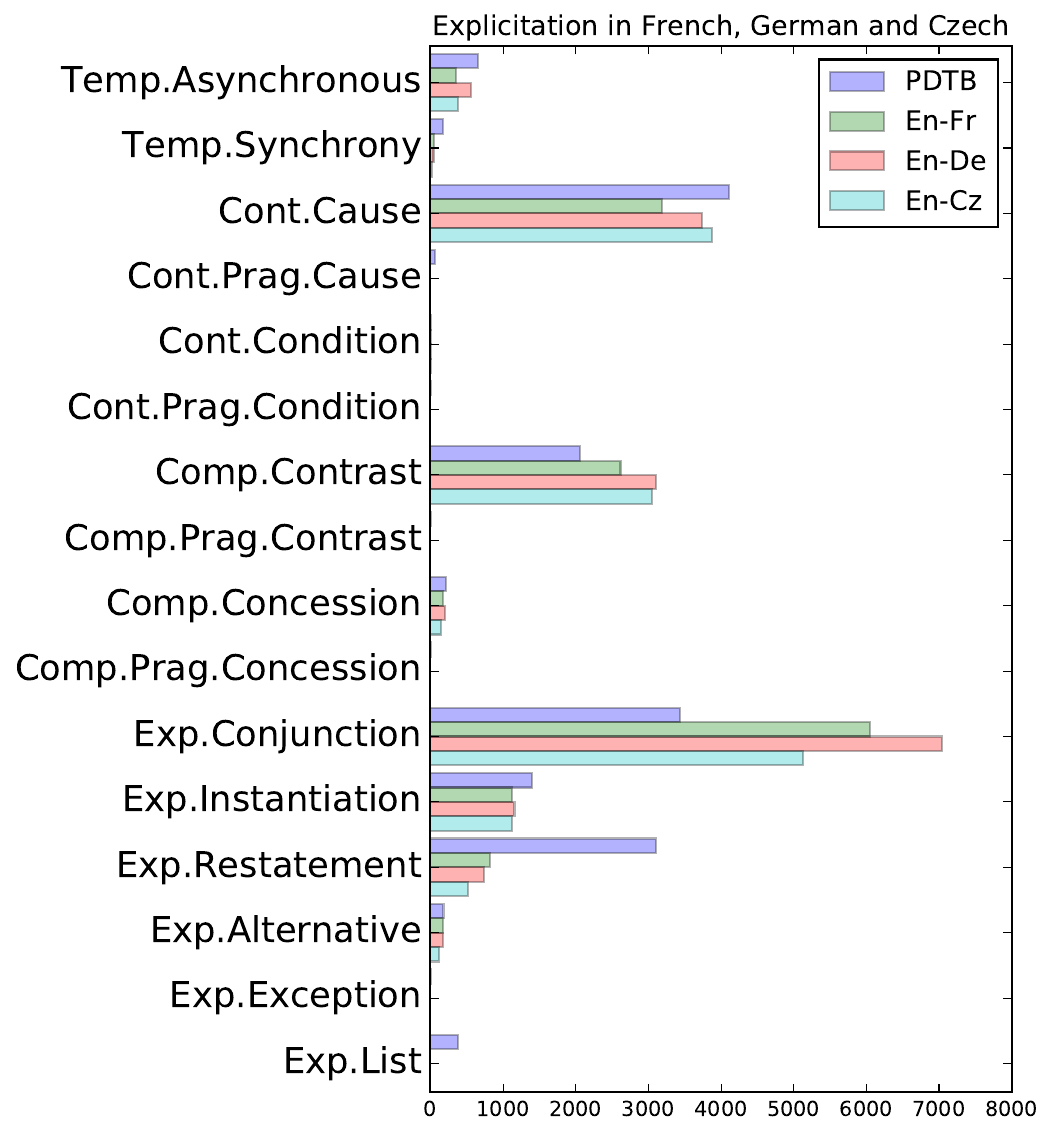}
    \caption{Distributions of PDTB and the extracted data among each discourse relation.}
    \label{fig:overall}
\end{figure}

\begin{figure*}[!htpb]
\centering
	\includegraphics[width=0.85\linewidth]{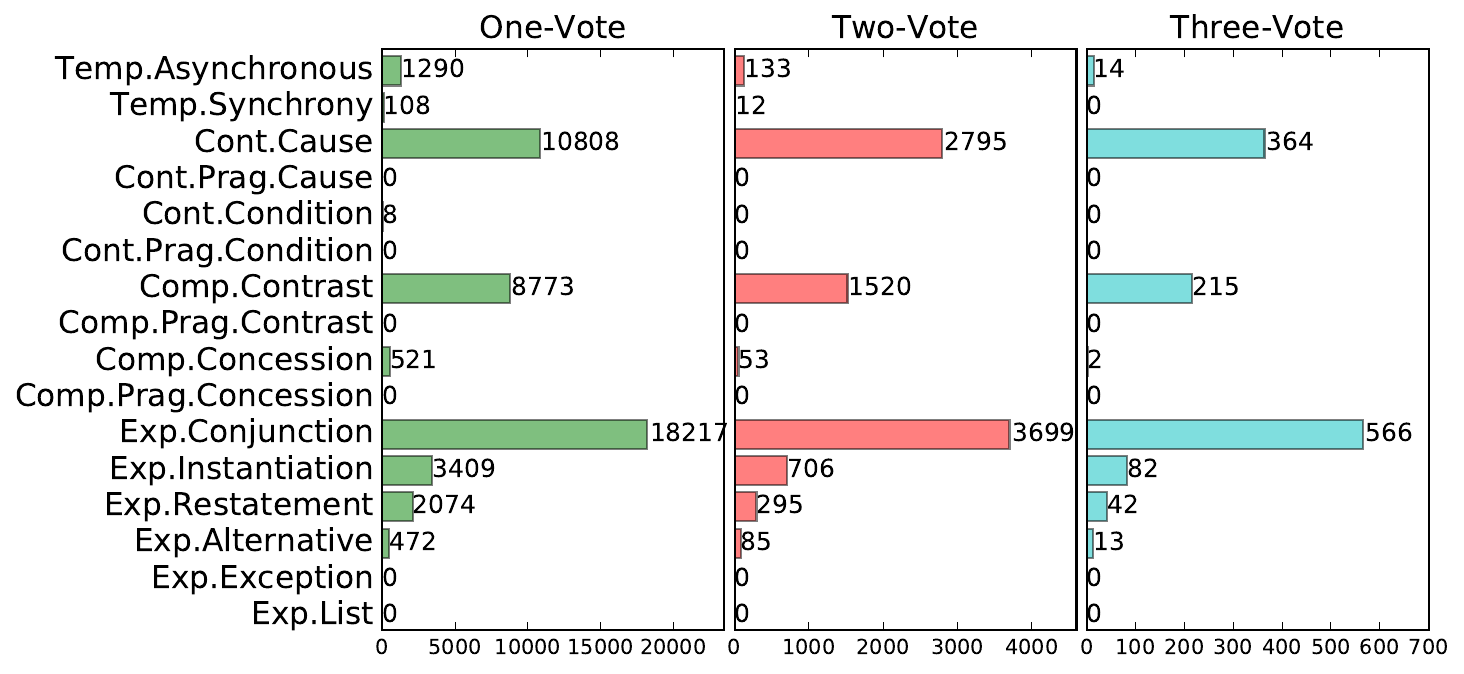}
    \caption{Distributions of discourse relations with different agreements.}
    \label{fig:votes}
\end{figure*}

\begin{table*}[!ht]
\centering
\begin{tabular}{clcccr}
\toprule
\multicolumn{2}{c}{} & PDTB-Lin & PDTB-Ji & Cross Validation & size of extra data \\ \midrule
\multicolumn{2}{l}{Majority Class} & 26.11 & 26.18 & 25.59 & - \\ 
\multicolumn{2}{l}{\citet{rutherford2017systematic}} & 38.38 & - & - & - \\
\multicolumn{2}{l}{\citet{shi2017using}} & \textbf{45.50} & - & 37.84 & 102,314 \\ 
\multicolumn{2}{l}{PDTB only}     & 37.95(0.59) & 40.57(0.67) & 37.82(0.14) & - \\ \midrule
\multirow{4}{*}{PDTB +}  & En-Fr  & 38.96(0.69) & 40.14(0.78) & 38.32(0.62) & 14,548 \\  
                          & En-De   & 39.65(0.95) & 39.96(0.44) & 37.97(0.46) & 16,757	\\ 
                          & En-Cz   & 37.90(1.27)  & 40.59(0.51) & 37.42(0.50)	& 14,375\\ 
                          & All     & 37.73(0.74) & 40.41(0.65) & 37.16(0.64)& 45,680  \\ \midrule
\multicolumn{2}{l}{PDTB + 2-votes} & \textbf{40.34}(0.75) & \textbf{41.95}(0.97) & \textbf{38.98}(0.14)& 9,298 \\ 
\multicolumn{2}{l}{PDTB + 3-votes} & 39.88(0.79) & 41.19(0.63) & 38.33(0.50)& 1,298 \\ 
\bottomrule
\end{tabular}
\caption{Performances with different sets of additional data. Average accuracy of 10 runs (5 for cross validations) are shown here with standard deviation in the brackets. Numbers in bold are significantly (p$<$0.05) better than the \textit{PDTB only} baseline with unpaired t-test.}
\label{table:results}
\end{table*}

\subsection{Results}
\subsubsection{Distribution of new instances}
Figure \ref{fig:overall} shows the distributions of expert-annotated PDTB implicit relations and the implicit discourse examples extracted from the French, German and Czech back-translations. Overall, there is no strong bias -- all relations seem to be represented similarly well, in line with their general frequency of occurrence. One interesting exception is the higher number of \textit{Expansion.Conjunction} relation from the German translations. The over-representation of \textit{Expansion.Conjunction} relation in German indicates that German translators tend to use more explicit cues to mark these relations. This is an independently discovered well-known finding from the literature \citep{kunz2015cross}, which observed that German tends to mark conjunction relations with discourse cues, while English tends to use coreference instead. We also find that  \textit{Expansion.Restatement} relations are under-represented in our back-translation method, indicating that these relations are explicitated particularly rarely in translation. 
We also find that we can identify more \textit{Contingency.Cause} and \textit{Comparison.Contrast} relations from the German and Czech back-translations compared to the French ones. This provides us with an interesting lead for future work, to investigate whether French tends to explicitate these relations less, expressing them implicitly like in the English original, or whether French connectives for causal and contrastive relations are more ambiguous, causing problems in the back-translations.

Figure \ref{fig:votes} shows that the filtering by majority votes (including only two cases where at least two back-translations agree with one another vs.~where all three agree) does again not change the distribution of extracted relations. 

In summary, we can conclude that the choice of translation language \textit{can} matter: depending on what types of relations are most important to acquire more data for the target task at hand, a language that tends to explicitate that relation frequently can be particularly suitable. 
On the other hand, if no strong such preferences on labelling specific relations exist, we can see that the choice of translation language only has a minor effect on the overall distribution of additional implicit discourse relation labels.

\subsubsection{Quantitative Results}

Table \ref{table:results} shows that best results are achieved by adding only those samples for which two back-translations agree with one another. This may represent the best trade-off between reliability of the label and the amount of additional data. The setting where the data from all languages is added performs badly despite the large number of samples, because this method contains different labels for the same argument pairs, for all those instances where the back-translations don't yield the same label, thus introducing noise into the system. The size of the extra data used in \citet{shi2017using} is about 10 times larger than our 2-votes data. The selection of instances differs in their paper from ours, in that they only use French, and in that they, unlike this paper, focus on intra-sentential samples. 
The model using the few reliable samples extracted from the back-translations of the three languages here significantly outperforms the results found in \citet{shi2017using} in the cross-validation setting. On the PDTB-Lin test set, we don't match performance, but note that this test set is based only on 800 instances, as opposed to the 16k instances in the cross-validation evaluation.

\subsubsection{Qualitative analysis}

Finally, we want to provide insight into what kind of instances the system extracts, and why back-translation labels sometimes disagree. We have identified four major cases based on a manual analysis of 100 randomly sampled instances.

\noindent\textbf{Case 1:} Sometimes, back-translations from several languages may yield the same connective because the original English sentence actually was not really unmarked, but rather contained an expression which could not be automatically recognized as a discourse relation marker by the automatic discourse parser\footnote{In the following examples, the original English sentence is shown is followed by the back-translations from French, German and Czech along with the connectives and senses.}. This can actually help us to identify new alternative lexicalisation for discourse relations, and thus represents a promising technique for improving discourse relation classification also on texts for which no translations are available.

\vspace{2pt}
\begin{small}
\noindent\textbf{Original English:}
I presided over a region crossed by heavy traffic from all over Europe, with significant accidents which gave rise to legal actions. \textbf{\textit{What is more,}} In 2002, two Member States of the European Union appealed to the European Court of Justice to repeal Directive 2002/15/EC because it included self-employed drivers ; the Court rejected their appeal on the grounds of road safety.\\
\noindent\textbf{French back-translation:}
I presided over a region crossed by heavy traffic from the whole of Europe, with significant accidents which gave rise to legal actions, \underline{\textbf{moreover}, (Expansion.Conjunction)} in 2002 , two Member States have appeal on the European Court of Justice, which has condemned the rejection of the grounds of road safety.\\
\noindent\textbf{German back-translation:} I presided over a region crossed by heavy traffic from across Europe, with significant accidents which, \underline{\textbf{moreover} (Expansion.Conjunction)} in 2002, two Member States of the European Union appealed to the European Court of Justice to repeal Directive 2002/15/EC , because it included self-employed drivers ; the Court quashed for reasons of road safety.\\
\noindent\textbf{Czech back-translation:} I was in the region with very heavy traffic from all over Europe, with significant accidents which gave rise to legal actions \underline{\textbf{therefore} (Contingency.Cause)} after all, in 2002, two Member States of the European Union appealed to the European Court of Justice to repeal Directive 2002/15/EC that also applies to self-employed drivers; the Court rejected their appeal on the grounds of road safety.\\
\end{small} \vspace{-6pt}

The expression \textit{what is more} is not part of the set of connectives labeled in PDTB and hence was not identified by the discourse parser. Our method is successful because such cues can be automatically identified from the consistent back-translations into two languages. (The case in Czech is more complex because the back-translation contains two signals, \textit{therefore} and \textit{after all}, see case 4.) 

We also found some similar expressions in this case like:

``in reality'' (``implicit'', original English) $\rightarrow$ ``in fact'' (explicit, back-translation); 

``for that reason'' $\rightarrow$ ``therefore''; 

``this is why'' $\rightarrow$ ``therefore''; 

``be that as it may'' $\rightarrow$ ``however / nevertheless''; 

``for another'' $\rightarrow$ ``furthermore / on the other hand''; 

``in spite of that'' $\rightarrow$ ``however / nevertheless'' and so on.

\vspace{6pt}
\noindent\textbf{Case 2:} Majority votes help to reduce noise related to errors introduced by the automatic pipeline, such as argument or connective misidentification: in the below example, \textit{also} in the French translation is actually the translation of \textit{along with}.

\vspace{2pt}
\begin{small}
\noindent\textbf{Original English:}
on behalf of the PPE-DE Group. (DE) Madam President, Commissioner, ladies and gentlemen, the public should be able to benefit in two ways from the potential for greater road safety. \textbf{\textit{For this reason}}, along with the report we are discussing today, I call for more research into the safety benefits of driver-assistance systems.\\
\noindent\textbf{French back-translation:}
(DE) Madam President, Commissioner, ladies and gentlemen, citizens should be able to benefit in two ways of the possibility of improving road safety. \underline{\textbf{also} (Expansion.Conjunction)} when we are discussing this report today, I appeal to the intensification of research at the level of the benefits of driver-assistance systems in terms of security, as well as the transmission of information about them.\\
\noindent\textbf{German back-translation:}
(DE) Madam President, Commissioner, ladies and gentlemen, road safety potentials should citizens in the dual sense \underline{\textbf{therefore} (Contingency.Cause)} I urge, together with the report under discussion today, the prevention and education about the safety benefits of driver-assistance systems.\\
\textbf{Czech back-translation:}
(DE) Madam President, Commissioner, ladies and gentlemen, the public would be the potential for greater road safety should have a two-fold benefit, \underline{\textbf{therefore} (Contingency.Cause)} I call, in addition to the report, which we are debating today , for more research and education in the safety benefits of driver-assistance systems.
\end{small}

\vspace{5pt}
\noindent \textbf{Case 3:} Discrepancies between connectives in back-translations can also be due to differences in how translators interpreted the original text. Here are cases of genuine ambiguities in the implicit discourse relation. 

\vspace{2pt}
\begin{small}
\noindent\textbf{Original English:} with regard, once again, to European Union law, we are dealing in this case with the domestic legal system of the Member States. \textbf{\textit{That being said}}, I cannot answer for the Council of Europe or for the European Court of Human Rights, which have issued a decision that I understand may raise some issues for Parliament.\\
\noindent\textbf{French back-translation:} with regard, once again, the right of the European Union, we are here in the domestic legal system of the Member States. \underline{\textbf{however,} (Comparison.Contrast)} I cannot respond to the place of the Council of Europe or for the European Court of Human Rights, which have issued a decision that I understand may raise questions in this House. \\
\noindent\textbf{German back-translation:} once again on the right of the European Union, we have it in this case with the national legal systems of the Member States. \underline{\textbf{therefore,} (Contingency.Cause)} I cannot, for the Council of Europe and the European Court of Human Rights, which have issued a decision, which I can understand, in Parliament raises some issues. \\
\noindent\textbf{Czech back-translation:} I repeat that, when it comes to the European Union, in this case we are dealing with the domestic legal system of the Member States. \underline{\textbf{in addition,} (Expansion.Conjunction)} I cannot answer for the Council of Europe or for the European Court of Human Rights , which has issued a decision that I understand may cause in Parliament some doubts. \\
\end{small}

\noindent\textbf{Case 4:} Implicit relations can co-occur with marked discourse relations \cite{rohde2015recovering}, and multiple translations help discover these instances, for example:

\vspace{2pt}
\begin{small}
\noindent\textbf{Original English:} We all understand that nobody can return Russia to the path of freedom and democracy,
\textbf{\textit{(implicit: but) what is more}}, the situation in our country is not as straightforward as it might appear to the superficial observer.\\
\noindent \textbf{French back-translation:} we all understand that nobody can return Russia on the path of freedom and democracy but Russia itself, its citizens and its civil society \underline{\textbf{but} (Comparison.Contrast)} there is more, the situation in our country is not as simple as it might appear to be a superficial observer.\\
\noindent\textbf{German back-translation:} we are all aware that nobody Russia back on the path of freedom and democracy, as the country itself, its people and its civil society \underline{\textbf{but} (Comparison.Contrast)} the situation in our country is not as straightforward as it might appear to the superficial observer. \\
\noindent\textbf{Czech back-translation:} we all know that Russia cannot return to the path of freedom and democracy there, but Russia itself, its people and civil society. \underline{\textbf{in addition} (Expansion.Conjunction)} the situation in our country is not as straightforward as it might appear to the superficial observer.
\end{small}

\section{Conclusion}
We compare the explicitations obtained from translations into three different languages, and find that instances where at least two back-translations agree yield the best quality, significantly outperforming a version of the model that does not use additional data, or uses data from just one language. 

We also found that specific properties of the translation language affect the distribution of the additionally acquired data across coherence relations: German, for instance, is known to mark conjunction relations using discourse cues more frequently, while English and other languages tend to express these relations rather through lexical cohesion or pronouns. This was reflected in our experiments: we found a larger proportion of explicitations for conjunction relations in German than the other translation languages. 

Finally, our qualitative analysis shows that the strength of the method partially stems from being able to learn additional discourse relation signals because these are typically translated consistently. The method thus shows promise for the identification of discourse markers and alternative lexicalisations, which can subsequently be exploited also for discourse relation classification in the absence of translation data. Our analysis also shows that our method is useful for identifying cases where multiple relations holding between two arguments.

\section{Acknowledgments}
We would like to thank all the anonymous reviewers for their careful reading and insightful comments. This research was funded by the German Research Foundation (DFG) as part of SFB 1102 ``Information Density and Linguistic Encoding''.

\bibliography{naaclhlt2019}
\bibliographystyle{acl_natbib}
\end{document}